\documentclass[journal]{IEEEtran}
\usepackage{times}
\usepackage{ifpdf}
\usepackage[pdftex]{graphicx}
\usepackage{amsmath}
\usepackage{amssymb}
\usepackage{subfigure}
\usepackage[ruled,commentsnumbered,noline]{algorithm2e}
\usepackage{paralist}
\usepackage{multirow}
\usepackage{slashbox}
\usepackage{cite}
\usepackage{algorithmic}
\usepackage{bm}
\usepackage{amsfonts}
\usepackage{graphicx}
\usepackage{textcomp}
\usepackage{CJK}
\usepackage{indentfirst}
\usepackage{makecell}
\usepackage{array}
\usepackage{float}
\usepackage{stfloats}
\usepackage{makecell}
\usepackage{amsthm,amsmath,amssymb}
\usepackage{mathrsfs}
\usepackage{color}
\usepackage{listings}
\usepackage{xcolor} %
\usepackage{pifont}  %

\begin{document}

\title{LatRef-Diff:  Latent and Reference-Guided Diffusion for Facial Attribute Editing and Style Manipulation}

\author{Wenmin~Huang, Weiqi~Luo, ~\IEEEmembership{Senior Member,~IEEE,} Xiaochun~Cao, ~\IEEEmembership{Senior Member,~IEEE,} Jiwu~Huang, ~\IEEEmembership{Fellow,~IEEE}

\thanks{This work was supported  in part by the National Natural Science Foundation of China under Grants  62472458. (Corresponding author: Weiqi~Luo.)}

\thanks{Wenmin Huang and Weiqi Luo are with GuangDong Province Key Lab of Information Security Technology and School of Computer Science and Engineering, Sun Yat-sen University, Guangdong 510000, China (E-mail: huangwm36@mail2.sysu.edu.cn; luoweiqi@mail.sysu.edu.cn).

Xiaochun Cao is with School of Cyber Science and Technology, Shenzhen Campus, Sun Yat-sen University, Shenzhen 518107, China (E-mail: caoxiaochun@mail.sysu.edu.cn).

Jiwu Huang is with the Guangdong Laboratory of Machine Perception and  Intelligent Computing, Faculty of Engineering, Shenzhen MSU-BIT University, Shenzhen, 518116, China (E-mail: jwhuang@smbu.edu.cn).}

}

\markboth{}%
{Shell \MakeLowercase{\textit{et al.}}: Bare Demo of IEEEtran.cls for IEEE Journals}

\maketitle

\begin{abstract}
Facial attribute editing and style manipulation are crucial for applications like virtual avatars and photo editing. However, achieving precise control over facial attributes without altering unrelated features is challenging due to the complexity of facial structures and the strong correlations between attributes. While conditional GANs have shown progress, they are limited by accuracy issues and training instability. Diffusion models, though promising, face challenges in style manipulation due to the limited expressiveness of semantic directions. In this paper, we propose LatRef-Diff, a novel diffusion-based framework that addresses these limitations. We replace the traditional semantic directions in diffusion models with style codes and propose two methods for generating them: latent and reference guidance. Based on these style codes, we design a style modulation module that integrates them into the target image, enabling both random and customized style manipulation. This module incorporates learnable vectors, cross-attention mechanisms, and a hierarchical design to improve accuracy and image quality. Additionally, to enhance training stability while eliminating the need for paired images (e.g., before and after editing), we propose a forward-backward consistency training strategy. This strategy first removes the target attribute approximately using image-specific semantic directions and then restores it via style modulation, guided by perceptual and classification losses. Extensive experiments on CelebA-HQ demonstrate that LatRef-Diff achieves state-of-the-art performance in both qualitative and quantitative evaluations. Ablation studies validate the effectiveness of our model's design choices.
\end{abstract}

\IEEEpeerreviewmaketitle
\section{Introduction}
\begin{figure}[!t]
\centering
\includegraphics[width=3.2in]{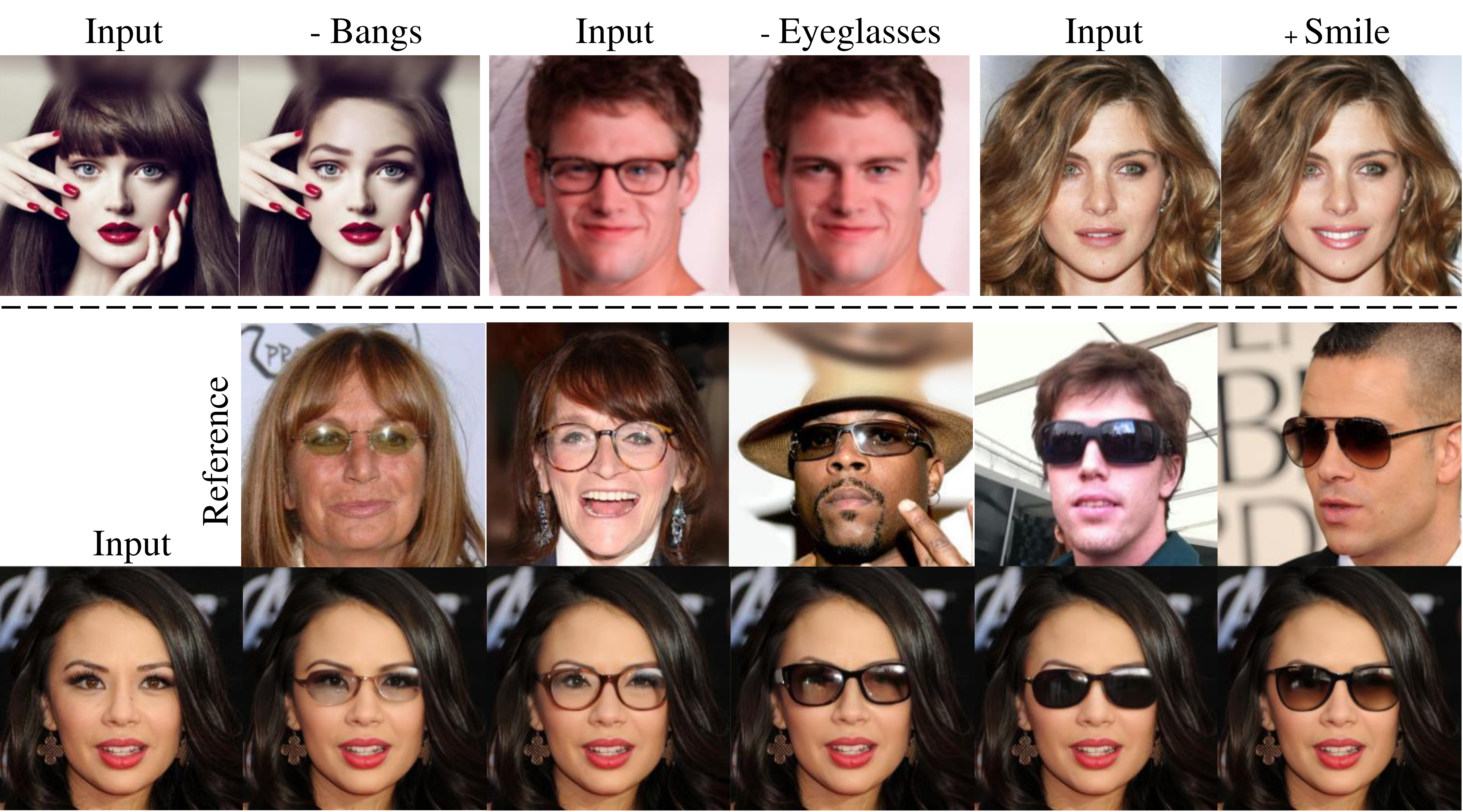}
\caption{Illustration of facial attribute editing (top) \& style manipulation (bottom) using the proposed LatRef-Diff.}
\label{fig_1}
\end{figure}

\IEEEPARstart{F}{acial} attribute editing commonly focuses on basic attribute-level modifications, such as adding or removing specific attributes~\cite{HeTIP2019,ChoiCVPR2018,LinCVPR2024} (top half of Fig.~\ref{fig_1}). In certain real-world applications, however, style manipulation goes beyond these simple modifications, offering more refined control over the appearance of attributes~\cite{DalvaTPAMI2023,HuangAAAI2024,10612246} (bottom half of Fig.~\ref{fig_1}). The applications of facial attribute editing and style manipulation are broad, ranging from virtual avatars and professional photo editing to data augmentation in facial recognition tasks. In such contexts, there is a strong demand for precise control, specifically the ability to modify a target attribute while preserving other facial features. Achieving this level of control is highly challenging, due to the inherent complexity of facial features and the strong correlations between attributes (e.g., age and eyeglasses).

In pursuit of achieving precise control, several works leveraging advanced conditional GANs~\cite{HearX2014} have emerged, focusing on two main  categories of methods: manipulating the latent space of pre-trained GANs and learning image-to-image translation. The first  kind of approach, exemplified by works such as~\cite{WangCVPR2022,PehlivanCVPR2023,9784910}, is  dedicated to attribute editing tasks. It begin by using GAN inversion to obtain the latent representation of a given face, then manipulates this representation along semantic directions and feeds it into a pre-trained GAN, such as the well-known StyleGAN series~\cite{KarrasCVPR2020}, to generate the edited result. The second kind of approach~\cite{HeTIP2019,HuangTCSVT2024}, which also primarily targets simple attribute editing, utilizes an encoder-decoder framework where attribute labels serve as conditioning inputs to guide the image editing process. It should be noted, however, that the aforementioned GAN-based methods are often limited in their ability to manipulate styles. To address this, a few advanced methods~\cite{DalvaTPAMI2023,HuangAAAI2024} adopt a more nuanced approach by manipulating styles through Gaussian noise or reference images for guidance, offering improved control over stylistic elements, such as distinguishing between myopic glasses and sunglasses. Despite significant advancements in GAN-based methods, their effectiveness in achieving high-quality and accurate attribute editing is often hindered by inherent GAN challenges, including non-convergence, training instability, and mode collapse~\cite{ArjovskyICLR2017, MeschederICML2018}.

Recently, diffusion models have demonstrated superior image synthesis quality compared to GANs and achieved remarkable success in various facial editing tasks, including text-guided identity-driven face generation \cite{RuizCVPR2024, 10769038}, stroke-based editing \cite{MengICLR2021}, and style transfer \cite{KimCVPR2022}. Motivated  by this progress, several recent studies on facial attribute editing \cite{PreechakulCVPR2022, KimCVPR2023} have shifted their attention to diffusion models. By adopting simplified versions of the variational lower bound~\cite{HoCVPR2020}, these methods optimize conditional DDIMs\cite{SongICLR2021} to learn a semantically rich latent space. Facial representations are then manipulated along semantic directions within this space to achieve editing effects. In contrast to GAN-based methods, which rely on unstable adversarial training, these diffusion-based approaches offer impressive attribute editing results through a more stable and straightforward self-reconstruction training scheme. However, a key limitation remains as the restricted expressiveness of semantic directions prevents these methods from performing style manipulation.

In this paper, we propose a novel Latent and Reference-guided Diffusion model (LatRef-Diff) for both facial attribute editing and style manipulation. Specifically, we  first  design style codes to replace the semantic directions used in previous diffusion models, removing the limitations on expressiveness. These style codes can be flexibly generated either by latent guidance from randomly sampled Gaussian noise or by reference guidance from a reference image.  Then, to achieve style manipulation, we propose a novel style modulation module that injects these style codes into the target image. Our style modulation module builds on AdaIN~\cite{HuangICCV2017} and incorporates innovations such as learnable vectors, cross-attention mechanisms, and a hierarchical design, achieving both precise and high-quality attribute editing and style manipulation.
Despite these advancements, training the model remains challenging. The self-reconstruction training scheme used in previous diffusion models is insufficient for effectively optimizing style codes and style modulation module.  While image pairs (i.e., original images and their corresponding edited versions) could help address this issue, they are often difficult to obtain in real applications. To address this challenge, we propose a forward-backward consistency training strategy inspired by the cycle consistency loss~\cite{ZhuCVPR2017}. This strategy first removes (forward) the target attribute approximately using image-specific semantic directions and then restores (backward) the attribute through style modulation. The proposed model is stably optimized in a non-adversarial manner by enforcing forward-backward consistency with perceptual and classification losses. The main contributions of this paper are threefold:

\begin{itemize}
\item We propose LatRef-Diff, which integrates style codes and  a style modulation module into the diffusion model, enabling both attribute editing and style manipulation with improved precise control.
\item We eliminate  the dependence on paired images by removing and restoring attributes, and propose a corresponding forward-backward consistency training strategy that effectively trains LatRef-Diff without adversarial loss.
\item Extensive experiments on CelebA-HQ~\cite{KarrasICLR2018} demonstrate that LatRef-Diff effectively tackles the style manipulation challenge in diffusion models, achieving state-of-the-art results in both qualitative and quantitative evaluations.
\end{itemize}

\section{Related Work}\label{sec2}

\subsection{Deep Generative Models}\label{sec2.1}
In recent years, various deep generative models have been proposed to generate samples across different modalities, such as text, audio, and images. In image generation, Generative Adversarial Networks (GANs)\cite{GoodfellowNIPS2014} have become one of the most popular models for tasks such as image inpainting, super-resolution, and image editing, owing to their ability to generate higher-quality samples compared to other models like Variational Autoencoders (VAEs)\cite{KingmaarX2013}, normalizing flows\cite{RezendearX2015}, and autoregressive models\cite{van2016pixel}. However, GANs often suffer from training instability and mode collapse, as their learning process depends on adversarial training between the generator and the discriminator to model the real data distribution\cite{ArjovskyICLR2017, MeschederICML2018}.
In contrast, diffusion models, such as DDPM\cite{HoNIPS2020} and DDIM\cite{SongICLR2021}, have recently demonstrated generative capabilities on par with GANs, without requiring adversarial training. These models train a denoising network on images corrupted by varying levels of Gaussian noise, allowing it to progressively remove the noise from the disturbed image. During inference, the model starts with a random sample of Gaussian noise and denoises it step by step into an image. With their ability to generate high-quality samples, stable training, and improved mode coverage, diffusion models have gained significant attention and emerged as powerful alternatives to GANs for image synthesis tasks.

\subsection{Facial Attribute Editing and Style Manipulation}\label{sec2.2}
\noindent \textbf{Facial Attribute Editing.}
Most existing facial attribute editing approaches are based on GANs. These methods have been extensively studied and can generally be categorized into two types:  manipulating the latent space of pre-trained GANs\cite{WangCVPR2022,PehlivanCVPR2023} and learning image-to-image translation\cite{GaoCVPR2021,HuangTCSVT2024,DalvaTPAMI2023,HuangAAAI2024}.
The first approach leverages the structured latent space of GAN models. Advanced GANs, particularly StyleGAN \cite{KarrasCVPR2020}, feature continuous and manipulable latent spaces where different directions correspond to each attribute\cite{KarrasCVPR2019,ShenTPAMI2020}. By performing GAN inversion, i.e., mapping facial images into the GAN's latent space, and shifting them along specific semantic directions, facial attribute editing can be achieved.
While the structured latent space of GANs provides an efficient mechanism for facial attribute editing, it is not without challenges.
The first challenge lies in the trade-off between reconstruction and editing. During GAN inversion, efforts to accurately preserve facial details often result in suboptimal embedding within the true distribution of the GAN's latent space.  This misalignment can render semantic directions less effective, thereby reducing editing ability.
The second challenge involves disentangling semantic directions.  The latent space semantic directions may not be fully independent, leading to unintended modifications of unrelated attributes during editing. To address these challenges, research efforts have focused on improving GAN inversion accuracy \cite{WangCVPR2022,PehlivanCVPR2023} and enhancing the disentanglement of semantic directions \cite{PatashnikCVPR2021,WuCVPR2021}, ensuring faithful and isolated attribute editing. The second category of methods typically treats facial attribute editing as an image-to-image translation task without paired images. Early approaches\cite{LiarX2016,ShenCVPR2017} divided the training data into two sets based on a specific attribute, such as eyeglasses: one set with eyeglasses and one set without. Then, an encoder-decoder network is trained to learn the translation between these two sets.
A major limitation of these methods is that a separate model must be trained for each attribute.
To improve efficiency, subsequent methods\cite{GaoCVPR2021,HuangTCSVT2024} introduced conditional GANs, using attribute labels as additional control conditions to guide the model in editing specific attributes, allowing a single model to support multiple attribute edits.

Recently, beyond GANs, diffusion models \cite{HoNIPS2020, SongICLR2021} have shown significant potential in various facial image manipulation tasks, such as text-guided identity-driven face generation \cite{RuizCVPR2024, 10769038}, stroke-based editing \cite{MengICLR2021}, and style transfer \cite{KimCVPR2022}.
Among these applications, we focus particularly on facial attribute editing tasks. Specifically, DiffAE \cite{PreechakulCVPR2022} introduces a learnable semantic encoder within DDIM \cite{SongICLR2021}, learning a semantically rich latent space in an autoencoding manner, similar to that of GANs.
DiffAE then edits images during the DDIM generative process by moving the latent codes along semantic directions defined by the weights of a linear classifier trained on latent codes of images with and without the target attribute.
Furthermore, DiffVAE \cite{KimCVPR2023} extends DiffAE to video-based facial attribute editing by replacing DiffAE's semantic encoder with an identity encoder and a landmark encoder, capturing identity and motion information to achieve temporal consistency across video frames.

\noindent \textbf{Style Manipulation.}
Although the aforementioned GAN-based and diffusion-based methods have made significant strides in facial attribute editing, they face limitations in style manipulation. This is primarily because the information contained in semantic directions and attribute labels is quite limited, which hinders these methods from capturing the full diversity of facial attributes. For example, faces with various types of eyeglasses, such as sunglasses, reading glasses, or fashion frames, can be classified under the ``with eyeglasses'' label, but they cannot be generated back from the same ``with eyeglasses'' label. As a result, these methods can only perform deterministic one-to-one mappings to simply add or remove target attributes, with no control over the style of the target attributes when added or removed.
To address this limitation, recent GAN-based image-to-image translation methods \cite{LiCVPR2021, HuangAAAI2024} have introduced style codes with strong expressive capabilities. These methods acquire the style code for an attribute either from randomly sampled Gaussian noise or from a reference image. By integrating the attribute label with the style code, they facilitate both attribute editing and style manipulation.  However, these methods rely on adversarial training, which can lead to instability and hinder their ability to generate high-accuracy, high-quality edits.

In this paper, we turn to diffusion models, known for their superior training stability, to tackle the aforementioned challenges. Specifically, inspired by style codes, we propose a novel style modulation module to address style manipulation in diffusion models. Furthermore, we introduce a new forward-backward consistency training strategy that effectively and stably optimizes the model, enabling both attribute editing and style manipulation.

\begin{figure*}[!t]
\centering
\includegraphics[width=7.0in]{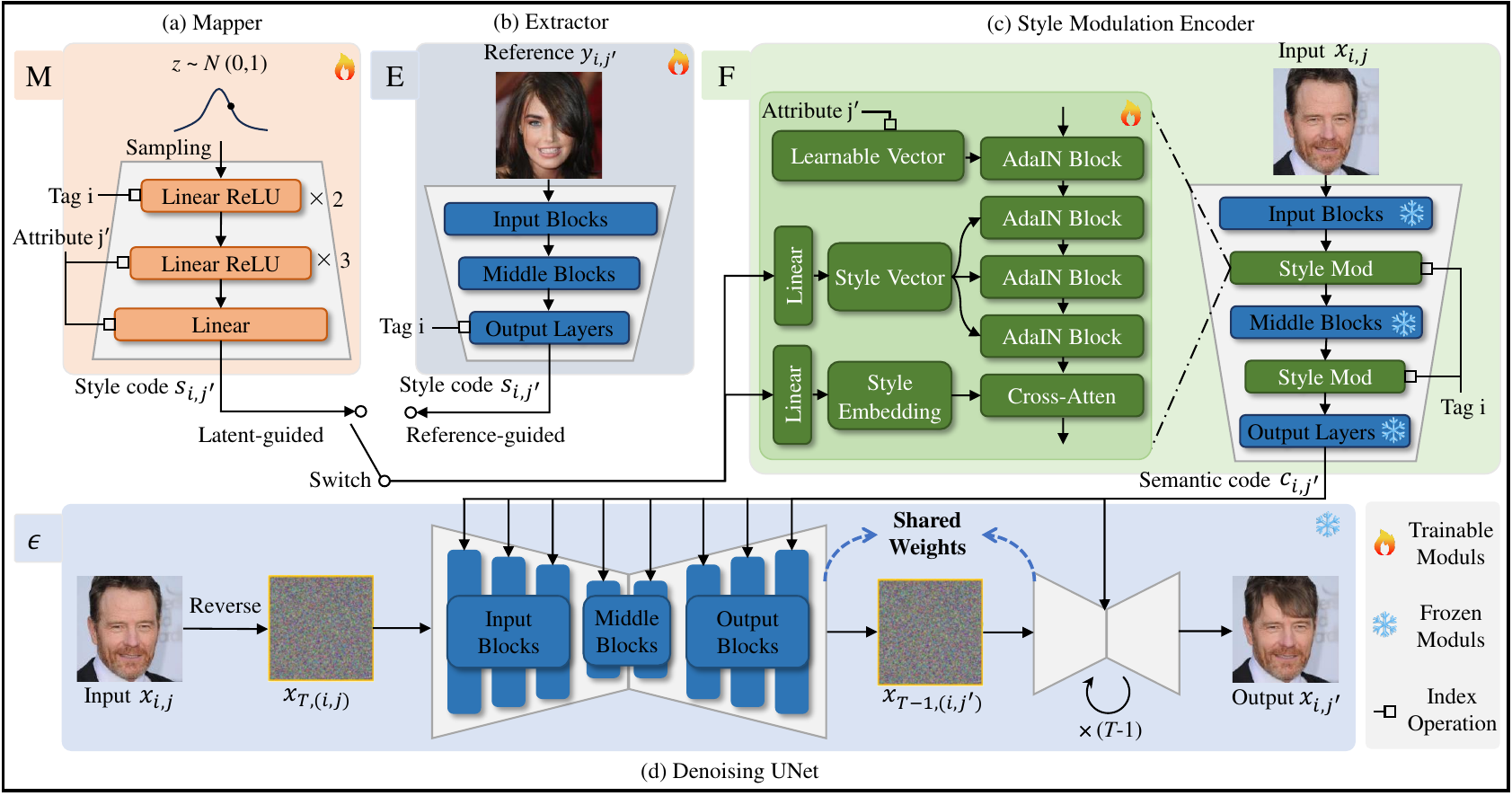}
\caption{The network architecture of the proposed LatRef-Diff, consisting of four components: (a) mapper $M$, (b) extractor $E$, (c) style modulation encoder $F$, and (d) denoising UNet $\epsilon$.}
\label{fig_2}
\end{figure*}

\section{Proposed Method}\label{sec3}
Following \cite{LiCVPR2021}, we use the subscript notation tag $i$ and attribute $j$ to denote an abstract attribute category and its corresponding specific attribute for an image (e.g., $x_{i,j}$ or $y_{i,j}$), or for image features (e.g., $s_{i,j}$ or $c_{i,j}$).  For example, when the tag $i$ represents ``bangs'', $j$ can denote specific attributes of the bangs, such as ``with bangs'' or ``without bangs''.

The primary goal of our LatRef-Diff is to address the limitations of diffusion-based methods in style manipulation and the instability of GAN-based methods during training, while effectively enhancing editing accuracy and image quality. To this end, we carefully design four key components in our method: a mapper $M$, an extractor $E$, a style modulation encoder $F$, and a denoising UNet $\epsilon$. As illustrated in Fig.~\ref{fig_2}, the method aims  to modify the specific attribute $j$ of an abstract attribute tag $i$ to a different attribute $j'$, obtaining the output image $x_{i,j'}$. To  modify the attribute $j$, we propose two implementation strategies: latent-guided and reference-guided, both generating the style code $s_{i,j'}$. In the latent-guided approach, the mapper $M$ generates a random attribute style code $s_{i,j'}$ from a noise sampled in the Gaussian distribution $\mathcal{N}(0, 1)$. In the reference-guided approach, the extractor $E$ extracts the attribute style code $s_{i,j'}$ from a given reference image $y_{i,j'}$. Subsequently, the resulting style code $s_{i,j'}$ and the input image $x_{i,j}$ are fed into the style modulation encoder $F$, which produces the semantic code $c_{i,j'}$. Finally, the denoising UNet $\epsilon$ iteratively denoises the initial latent noise $x_{T,(i,j)}$ of input image $x_{i,j}$, conditioned on $c_{i,j'}$, to generate the output edited image $x_{i,j'}$.

In the following sections, we provide a detailed description of each component within LatRef-Diff and discuss how they are effectively optimized using the proposed training strategy.

\subsection{Mapper and Extractor}
\label{sec3.1}
Based on the specific requirements of the task, our method provides two ways to obtain style codes $s_{i,j'}$: latent-guided and reference-guided. These two approaches are implemented by the following modules:

\noindent \textbf{Mapper.} The mapper $M$ consists of a multilayer perceptron (MLP). To minimize interference between different attributes, the first two LinearReLU layers of the MLP index using tag $i$, while the subsequent four LinearReLU layers index using attribute $j'$, generating the style code. Formally, given a  Gaussian noise $z$, and attribute $j'$ for tag $i$, $M$ generates a style code $s_{i,j'} = M_{i,j'}(z)$.

\noindent \textbf{Extractor.} The extractor $E$  first extracts facial features through the input blocks and middle blocks, as defined in~\cite{DhariwalNIPS2021}, which include downsampling operations. It then indexes the style code from the output layers using tag $i$. Formally, given a reference image $y_{i,j'}$, $E$ extracts a style code $s_{i,j'} = E_i(y_{i,j'})$.

\subsection{Style Modulation Encoder}\label{sec3.2}
Existing diffusion-based methods edit attributes along predefined semantic directions, which struggle with fine-grained style manipulation. In this paper, we propose a style modulation encoder $F$ to replace semantic directions for attribute manipulation, addressing the challenge of style manipulation in diffusion models.
As illustrated in Fig.~\ref{fig_2}(c), $F$ consists of four components: input blocks, style modulation module, middle blocks, and output layers. The input block comprises a series of convolutional layers with downsampling operations, which transform raw pixel data into high-dimensional abstract representations by reducing the spatial resolution of feature maps.
This step allows $F$ to effectively capture both content and attribute information from facial images, which are then passed to the style modulation module.

The style modulation module serves as the core of $F$,  enabling precise facial attribute editing and style manipulation based on style codes. Previous GAN-based methods typically use AdaIN~\cite{HuangICCV2017} to adjust the mean and variance of facial features conditioned on style codes for style modulation. However, this simple approach is highly dependent on the accuracy of style code extraction, lacks effective interaction between local and global information, and is prone to interference between attributes, which significantly impacts the model's attribute manipulation ability and the quality of editing images. For more details, refer to Section IV-E-1.
To address these challenges, we propose three key improvements for style modulation.
\begin{enumerate}
\item\textbf{Learnable Vector}:
We introduce an additional learnable vector as the conditional input for AdaIN. It is learned across the entire dataset and is not influenced by specific Gaussian noise, reference images, or style extraction networks like $M$ or $E$, thereby improving the reliability of attribute editing and style manipulation.

\item \textbf{Cross-Attention Mechanism}: After AdaIN modulation, we introduce cross-attention mechanism to enhances the interaction between local and global information, thereby improving the overall quality of image editing.

\item \textbf{Hierarchical Design}: The style modulation adopts a hierarchical design of tags $\rightarrow$ attributes. Specifically, each tag is assigned a dedicated style modulation unit, within which different learnable vectors are provided for the various attributes associated with that tag. This hierarchical approach effectively minimizes interference between attributes.

\end{enumerate}

The middle blocks and output layer serve as transition layers, interleaved between style modulation module, to further refine the modulated features and generate the final facial semantic  code. Formally, given a facial image $x_{i,j}$ and a style code $s_{i,j'}$, the style modulation encoder $F$ edits the attribute $j$ to $j'$, and outputs the semantic code $c_{i,j'}=F_{i,j'}(x_{i,j},s_{i,j'})$.

\subsection{Denoising UNet}\label{sec3.3}
Diffusion models\cite{HoNIPS2020,SongICLR2021} are a family of generative models that  learn to model the target distribution by training a denoising UNet to reverse a noising process at various noise levels. Notably, DDIM within diffusion models demonstrates strong reversibility in image encoding and decoding, making it well-suited for image editing tasks.
In this work, we utilize a variant of DDIM, known as conditional DDIM (cDDIM) \cite{PreechakulCVPR2022}, for facial attribute editing.
cDDIM consists of a noising forward process and a denoising generative process. Specifically, we follow the non-Markovian forward process $q(x_{t-1} | x_t, x_0)$ in DDIM to add noise to the input image $x_0$. The noisy version of  $x_0$ at time $t$ (out of $T$), $x_t$, can be expressed as $x_t=\sqrt{\alpha_t}x_0+\sqrt{1-\alpha_t}\epsilon_t$, $\epsilon_t \sim\mathcal{N}(0, \mathbf{I})$
where $\alpha_t =\prod_{s=1}^t(1-\beta_s)$, and  $\beta_s$ is a hyperparameter controlling the noise levels.
Our goal is to learn the distribution $p_\theta(x_{t-1} | x_t, c_{i,j})$ to approximate $q(x_{t-1} | x_t, x_0)$. When the difference between $t$ and $t-1$ is infinitesimally small, $p_\theta(x_{t-1} | x_t, c_{i,j})$ can be modeled as $p_\theta(x_{t-1} | x_t, c_{i,j}) = \mathcal{N}(\mu_\theta (x_t,t,c_{i,j}), \mathbf{0})$\cite{MeschederICML2015,SongICLR2021}. We employ a denoising UNet $\epsilon_\theta$ to predict the noise \( \epsilon \) for reparameterizing \( \mu_\theta \):
\begin{equation}\label{eq1}
\begin{aligned}
\mu_{\theta}(x_t,t,c_{i,j}) = &\ \sqrt{\alpha_{t-1}} \left( \frac{x_t - \sqrt{1 - \alpha_t}\epsilon_\theta(x_t, t,c_{i,j})}{\sqrt{\alpha_t}} \right) \\
&+ \sqrt{1 - \alpha_{t-1}}\epsilon_\theta(x_t, t,c_{i,j}).
\end{aligned}
\end{equation}
The optimization objective is given as $\mathbb{E}_{x_0,\epsilon,t}=||\epsilon_\theta(x_t, t,c_{i,j})-\epsilon||_2^2$. Using this, our cDDIM follows a deterministic generation process by iterating $x_{t-1} = \mu_{\theta}(x_t,t,c_{i,j})$, and its reverse process can be given by:
\begin{equation}\label{eq2}
\begin{aligned}
x_{t+1} = &\  \sqrt{\alpha_{t+1}} \left( \frac{x_t - \sqrt{1 - \alpha_t} \, \epsilon_\theta(x_t, t,c_{i,j})}{\sqrt{\alpha_t}} \right) \\
&+ \sqrt{1 - \alpha_{t+1}} \, \epsilon_\theta(x_t, t,c_{i,j}).
\end{aligned}
\end{equation}
This reversibility allows us to encode and decode facial images. Formally, given an image $x_{i,j}$ to be edited and its corresponding semantic code $c_{i,j}$, we first run the reverse process of deterministic generative process to obtain the initial latent noise:
\begin{equation}\label{eq3}
\begin{aligned}
x_{T,(i,j)} = \text{cDDIM}_{\text{enc}}(\epsilon_\theta; x_{i,j}, c_{i,j}),
\end{aligned}
\end{equation}
where $x_{T,(i,j)}$ is intended to encode only the information left out by $c_{i,j}$, i.e., the stochastic details. Next, conditioned on a new semantic code $ c_{i,j'}$, we decode $x_{T,(i,j)}$ in the generative process to obtain the editing image:
\begin{equation}\label{eq4}
\begin{aligned}
x_{i,j'} = \text{cDDIM}_{\text{dec}}(\epsilon_\theta; x_{T,(i,j)}, c_{i,j'}).
\end{aligned}
\end{equation}
This generative process enables $x_{i,j'}$ to adhere to the true data distribution, ensuring the quality of the editing image. The detailed encoding and decoding process for facial images is outlined in Algorithm \ref{Alg1}.
\begin{algorithm}[t]
\caption{LatRef-Diff}\label{Alg1}
\KwIn{$E$, $M$, $F$, $\epsilon_{\theta}$, $x_{i,j}$,$y_{i,j'}$}
\KwOut{editing image $x_{i,j'}$}
\nlset{1}$s_{i,j} = E_i(x_{i,j})$
\footnotesize
\tcp*[f]{\color{gray} Image Encoding:1-7}\\
\normalsize
\nlset{2}$c_{i,j} = F_{i,j}(x_{i,j},s_{i,j})$\\
\nlset{3}$x_{0,(i,j)} \leftarrow x_{i,j}$\\
\SetInd{0.5em}{0.3em}
\nlset{4}\textbf{for} $t = 0, \dots, T-1$ \textbf{do}
\footnotesize
\tcp*[f]{\color{gray} $x_{T,(i,j)} = \text{cDDIM}_{\text{enc}}(\epsilon_\theta; x_{i,j}, c_{i,j})$}\\
\normalsize
\nlset{5}\footnotesize
\hspace{1em}$x_{t+1,(i,j)} =   \sqrt{\alpha_{t+1}} \left( \frac{x_{t,(i,j)} - \sqrt{1 - \alpha_t} \, \epsilon_\theta(x_{t,(i,j)}, t,c_{i,j})}{\sqrt{\alpha_t}} \right) + $\\
 $\hspace{7.6em}\sqrt{1 - \alpha_{t+1}} \, \epsilon_\theta(x_{t,(i,j)}, t,c_{i,j}) $

\normalsize
\nlset{6}\textbf{end for}\\
\nlset{7}    $x_{T,(i,j)} \leftarrow x_{t+1,(i,j)}$\\ 
\nlset{8} \textbf{switch} (Guided Manner):
\footnotesize
\tcp*[f]{\color{gray} Image Decoding:8-18}\\
\normalsize
\nlset{9} \hspace{1.6em}\textbf{case 1:} Reference-guided \\
\nlset{10}\hspace{5em}$s_{i,j'} = E_i(y_{i,j'})$\\
\nlset{11}\hspace{1.6em}\textbf{case 2:} Latent-guided \\
\nlset{12}\hspace{5em}$s_{i,j'}=M_{i,j'}(z),$ $z\sim\mathcal{N}(0, \mathbf{I})$\\
\nlset{13}\textbf{end switch}

\nlset{14}$c_{i,j'} = F_{i,j'}(x_{i,j},s_{i,j'})$\\

\SetInd{0.5em}{0.3em}
\nlset{15}\textbf{for} $t = T, \dots, 1$ \textbf{do}
\footnotesize
\tcp*[f]{\color{gray} $x_{i,j'} = \text{cDDIM}_{\text{dec}}(\epsilon_\theta; x_{T,(i,j)}, c_{i,j'})$}\\
\normalsize
\nlset{16}\footnotesize\hspace{1em}$x_{t-1,(i,j')}=\sqrt{\alpha_{t-1}} \left( \frac{x_{t,(i,j')} - \sqrt{1 - \alpha_t}\epsilon_\theta(x_{t,(i,j')}, t,c_{i,j'})}{\sqrt{\alpha_t}} \right) +$\\ $\hspace{8em} \sqrt{1 - \alpha_{t-1}}\epsilon_\theta(x_{t,(i,j')}, t,c_{i,j'})$\\
\normalsize
\nlset{17}\textbf{end for}\\
\nlset{18}    $x_{i,j'} \leftarrow x_{t-1,(i,j')}$\\ 
\nlset{19}\Return $x_{i,j'}$ 


\end{algorithm}

\subsection{Training Objectives}\label{sec3.4}
\begin{figure}[!t]
\centering
\includegraphics[width=3.5in]{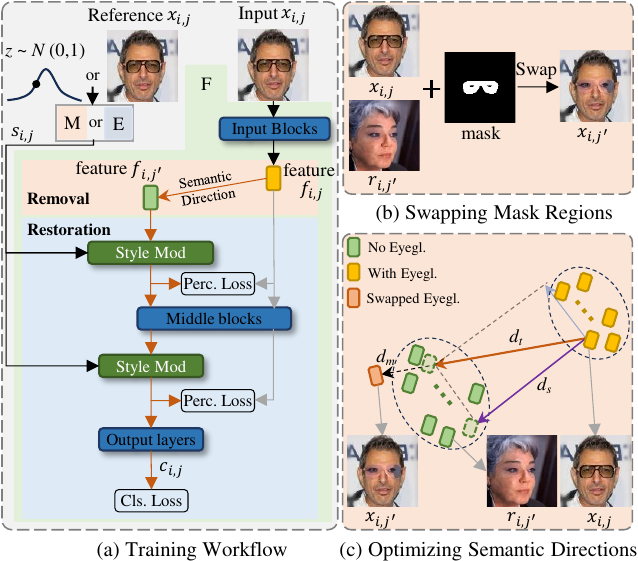}
\caption{Detailed workflow of the forward-backward consistency training strategy for LatRef-Diff.}
\label{fig_3}
\end{figure}

Our goal is to train LatRef-Diff to enable attribute editing and style manipulation using either latent guidance or reference guidance.
A straightforward approach is to use paired images before and after editing, such as $x_{i,j}$ and $x_{i,j'}$, and then optimize the model using the loss function $||\epsilon_\theta(x_{t,(i,j)}, t,c_{i,j})-\epsilon||_2^2$, where $c_{i,j}=F_{i,j}(x_{i,j'},E_i(x_{i,j}))$. However, paired images with different attributes for the same identity are typically unavailable.
Simply replacing $x_{i,j'}$ with $x_{i,j}$ to extract $c_{i,j}$ does not guarantee that the style modulation module performs attribute editing as expected. This is because the $F$ may ignore the style code $s_{i,j}$ and learn a trivial mapping, such as $c_{i,j} = F_{i,j}(x_{i,j},\cdot)$. For more details, refer to the ``w/ SVLB'' in Table~\ref{Table 4}.
To address this issue, we propose a novel forward-backward consistency training strategy, as shown in Fig.~\ref{fig_3}(a).
The key idea is to first approximately remove the target attribute in the latent space (for example, changing $j$ to $j'$), and then restore the target attribute through style modulation, editing $j'$ back to $j$. This approach prevent $F$ from learning trivial mappings. The details of the removing and restoring processes are as follows.

\noindent \textbf{Removal.} Previous studies\cite{PreechakulCVPR2022} have shown that a well-trained semantic encoder organizes its latent space with interpretable directions, where moving along specific directions modifies specific facial attributes.
One straightforward approach to identifying the direction of a given attribute is to locate sets of images with and without the target attribute in the latent space, compute the means of these sets, and then take their difference. For instance, the semantic direction of removing eyeglasses, as shown in Fig.~\ref{fig_3}(c), can be expressed as:
\begin{equation}\label{eq5}
\begin{small}
\begin{aligned}
d_{\text{s}} = \frac{1}{|D_{i,j'}|}\!\sum_{y_{i,j'}\in D_{i,j'}}\!\!\!\!\!\!\phi(y_{i,j'}) - \frac{1}{|D_{i,j'}|}\!\sum_{x_{i,j}\in D_{i,j}}\!\!\!\!\!\phi(x_{i,j}),
\end{aligned}
\end{small}
\end{equation}
where $D_{i,j'}$ and $D_{i,j}$ represent the sets of images without eyeglasses and with eyeglasses, respectively. $|\cdot|$ denotes the cardinality of the set, and $\phi$ represents the Input Blocks.
With sufficient positive and negative samples for eyeglasses, $d_{s}$ reliably removing the eyeglasses. However, due to inter-subject variability, uniformly applying  $d_{s}$ to all images may not ensure that non-eyeglasses attributes and other details remain unaffected during the eyeglasses removal process.
To address this issue, we propose a method that adaptively adjusts $d_{s}$ for each image to generate an image-specific semantic direction $d_t$.
Specifically, for a given face image $x_{i,j}$ with eyeglasses, we first obtain a semantic mask indicating eyeglasses region. Using this mask, we swap the eyeglasses region between $x_{i,j}$ and a randomly selected face image $r_{i,j'}$ without eyeglasses, resulting in a new image $x_{i,j'}$, as illustrated in Fig.~\ref{fig_3}(b). This allows us to compute a direction  $d_m=\phi(x_{i,j'})-\phi(x_{i,j})$.
Given the interpretability of the latent space, we hypothesize that moving along $d_m$ primarily affects eyeglasses regions.
To refine $d_m$ and ensure effective attribute editing, we adjust its length based on $d_{s}$ to obtain the final direction $d_{t}$.  Specifically, we require that one component of $d_t$, when orthogonally decomposed, equals $d_s$. This relationship is defined as:
\begin{equation}\label{eq6}
\begin{aligned}
d_{t} = \frac{||d_s||^2}{d_m \cdot d_s} \cdot d_m.
\end{aligned}
\end{equation}
By moving $f_{i,j}$ along $d_t$, we can approximately remove eyeglasses while preserving other content as much as possible.

\vspace{0.5em} \noindent \textbf{{Restoration}.}  In this stage, our objective is to use the style modulation module to edit the modified feature $f_{i,j'}$ back to its original state $f_{i,j}$, effectively restoring it to its state before modification. To achieve this, we introduce the perceptual loss, commonly used in image reconstruction and style transfer.
In these tasks, perceptual loss minimizes the distance between feature maps extracted from a pre-trained network, such as VGG\cite{SimonyanarX2014}, for the original and generated images, thereby enhancing the similarity in detail and perception.
Inspired by this, we freeze all parameters in $F$ except the style modulation module $SM$ and minimize the discrepancy between $SM(f_{i,j'},s_{i,j})$ and $f_{i,j}$.
This encourages the style modulation module to restore $f_{i,j}$ from $f_{i,j'}$. Our perceptual loss is defined as:
\begin{equation}\label{eq7}
\begin{aligned}
\mathcal{L}_{perc.} = ||SM(f_{i,j'},s_{i,j})-f_{i,j}||^2_2.
\end{aligned}
\end{equation}
Although perceptual loss captures the overall content of the image well, our experiments revealed that it lacks sensitivity to specific attribute information.  Therefore, we also introduce a classification loss to further constrain the final output $c_{ij}$ of $F$. The classification loss is defined as:
\begin{equation}\label{eq8}
\begin{small}
\begin{aligned}
\mathcal{L}_{cls.}=-l^\top\log{}C(c_{i,j})-(1-l^\top)\log{}(1-C(c_{i,j})),
\end{aligned}
\end{small}
\end{equation}
where $l$$\in$$\mathbb{R}$$^{n}$ is the $n$-dimensional attribute value vector for $x_{i,j}$, $\top$ indicates the transposition operation, and $C(\cdot)$$\in$$\mathbb{R}$$^{n}$ indicates the prediction of $n$ attribute value. Finally, our full objective function is written as follows:
\begin{equation}\label{eq18}
\begin{aligned}
\mathcal{L}_{full}=\mathcal{L}_{perc.}+\mathcal{L}_{cls.}.
\end{aligned}
\end{equation}

\begin{figure*}[!ht]
\centering
\includegraphics[width=7.0in]{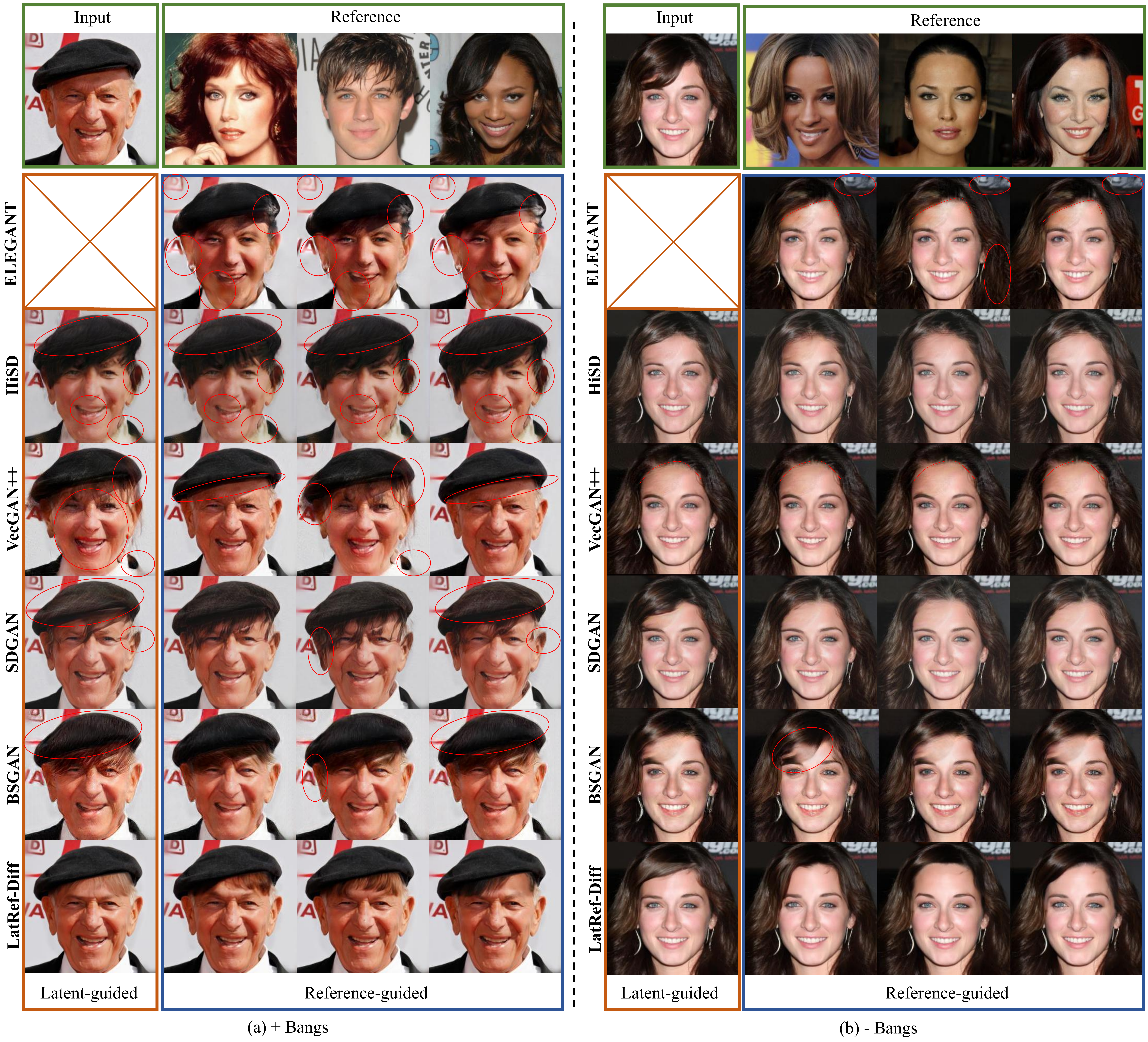}
\caption{Qualitative results of style manipulation under setting \#1, including (a) + Bangs and (b) - Bangs. In the following figures, we highlight the problematic areas in the generated images that require magnification to observe the detailed differences.  }
\label{fig_4}
\end{figure*}

\begin{figure*}[!ht]
\centering
\includegraphics[width=7.1in]{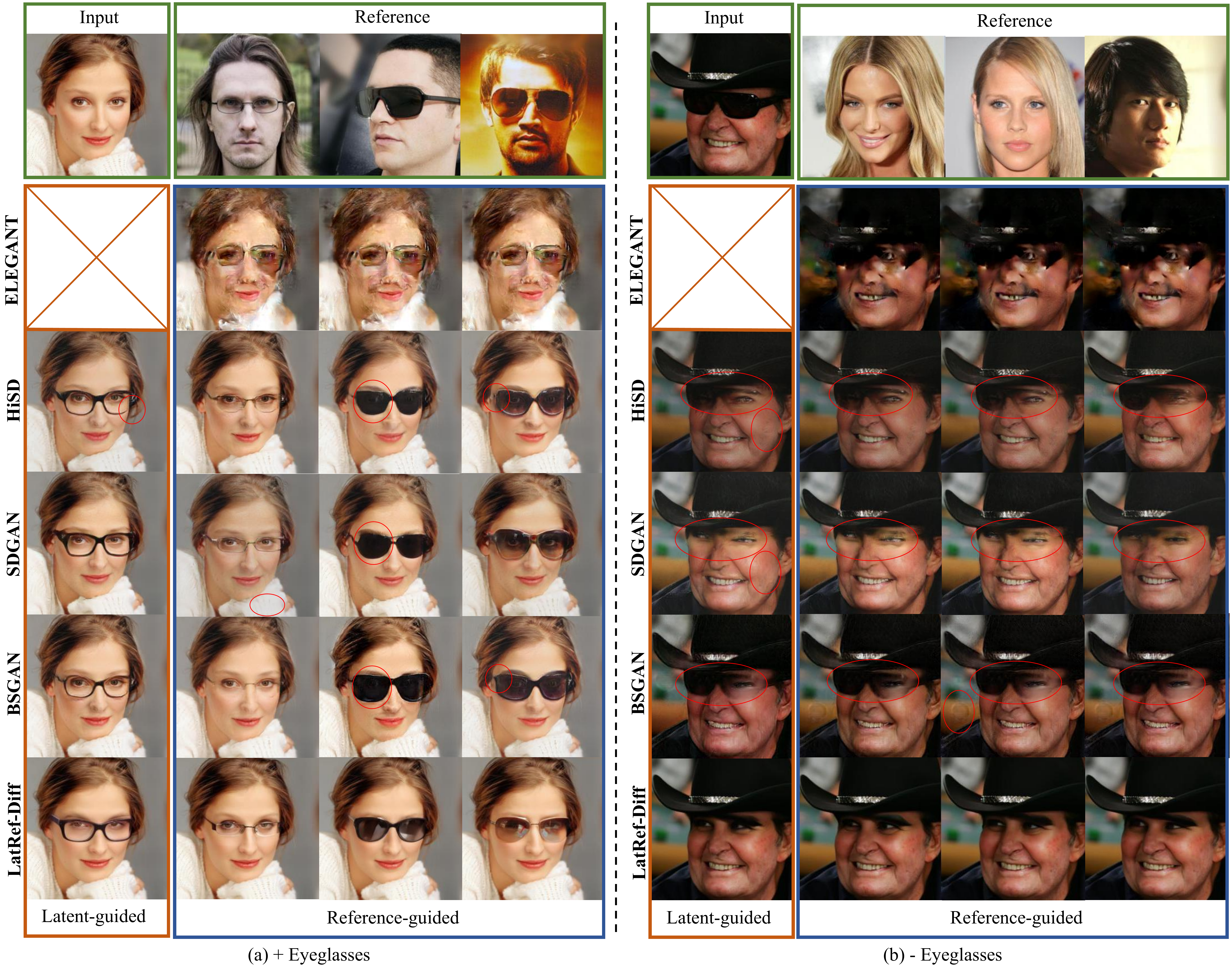}
\caption{Qualitative results of style manipulation under setting \#1, including (a) + Eyeglasses and (b) - Eyeglasses.}
\label{fig_5}
\end{figure*}

\section{Experiments}\label{sec4}

\subsection{Database and Setting}\label{sec4.1}
As in previous studies \cite{LiCVPR2021,DalvaTPAMI2023,HuangAAAI2024}, we use the CelebA-HQ database~\cite{KarrasICLR2018} for our experiments. This database contains $30,000$ high-quality facial images with simple binary attribute annotations, yet it lacks any style-related information. Since our method is capable of both style manipulation and attribute editing, which require different experimental setups, we will present them separately below:

\begin{table}[t]
\caption{Comparison of Our Method and State-of-the-Art in Style Manipulation and Attribute Editing}\label{Table 0}
      \centering
      \makegapedcells
      \setcellgapes{2pt}
      \setlength\tabcolsep{4pt}
      \begin{tabular}{|l || c | c | c | c|}
        \hline
        \multirow{2}{*}{Methods} & \multicolumn{3}{c|}{Style Manipulation} & \multicolumn{1}{c|}{Attribute} \\
        \cline{2-4}
        & Latent & Reference & Text & Editing \\
        \hline
        AttGAN (TIP 2019)\cite{HeTIP2019} & \ding{55} & \ding{55} & \ding{55} & \ding{51} \\
        \hline
        HFGI (CVPR 2022)\cite{WangCVPR2022} & \ding{55} & \ding{55} & \ding{55} & \ding{51} \\
        \hline
        DiffAE (CVPR 2022)\cite{PreechakulCVPR2022} & \ding{55} & \ding{55} & \ding{55} & \ding{51} \\
        \hline
        StyleRes (CVPR 2023)\cite{PehlivanCVPR2023} & \ding{55} & \ding{55} & \ding{55} & \ding{51} \\
        \hline
        ELEGANT (ECCV 2018)\cite{XiaoECCV2018} & \ding{55} & \ding{51} & \ding{55} & \ding{51} \\
        \hline
        HiSD (CVPR 2021)\cite{LiCVPR2021} & \ding{51} & \ding{51} & \ding{55} & \ding{51} \\
        \hline
        VecGAN++ (TPAMI 2023)\cite{DalvaTPAMI2023} & \ding{51} & \ding{51} & \ding{55} & \ding{51} \\
        \hline
        SDGAN (AAAI 2024)\cite{HuangAAAI2024} & \ding{51} & \ding{51} & \ding{55} & \ding{51} \\
        \hline
        BSGAN (ESWA 2025)\cite{ren2025facial} & \ding{51} & \ding{51} & \ding{55} & \ding{51} \\
        \hline
        IP2P (CVPR 2023)\cite{brooks2023instructpix2pix} & \ding{55} & \ding{55} & \ding{51} & \ding{51} \\
        \hline
        I-CLIP (CVPR 2025)\cite{chen2025instruct} & \ding{55} & \ding{55} & \ding{51} & \ding{51} \\
        \hline
        LatRef-Diff & \ding{51} & \ding{51} & \ding{55} & \ding{51} \\
        \hline
      \end{tabular}
\end{table}

\noindent \textbf{Setting \#1: Style Manipulation.}
Following the protocol of HiSD~\cite{LiCVPR2021}, we use $27,000$ images from the CelebA-HQ database for training and reserve the remaining $3,000$ images for testing. The attributes \textit{Bangs} and \textit{Eyeglasses} are selected for evaluation due to their distinct style differences and their frequent use in related works. To facilitate comparative analysis, we include five  methods that support attribute style manipulation: ELEGANT~\cite{XiaoECCV2018}, HiSD~\cite{LiCVPR2021}, VecGAN++~\cite{DalvaTPAMI2023}, SDGAN~\cite{HuangAAAI2024} and BSGAN~\cite{ren2025facial}, as shown in Table~\ref{Table 0}.

Beyond face-attribute editing baselines, we further broaden the comparison to the recently emerging text-to-image foundation models, which are trained on billions of image-text pairs and demonstrate strong generalization. From this category, we select IP2P\cite{brooks2023instructpix2pix} and I-CLIP\cite{chen2025instruct}, as their ability to perform flexible image editing through textual instructions, together with their strong preservation of irrelevant content, enables them to  meet the requirements of face attribute editing.

\noindent \textbf{Setting \#2: Attribute Editing.}
Following the setup of VecGAN++~\cite{DalvaTPAMI2023}, we use $27,176$ images from the CelebA-HQ dataset for training and the remaining $2,824$ images for testing. Three attributes, \textit{Smile}, \textit{Young}, and \textit{Male}, which are commonly used in StyleGAN inversion methods, are selected for evaluation. To facilitate comparative analysis, six state-of-the-art methods that support attribute editing are included: AttGAN~\cite{HeTIP2019}, HFGI~\cite{WangCVPR2022}, StyleRes~\cite{PehlivanCVPR2023}, DiffAE~\cite{PreechakulCVPR2022}, SDGAN~\cite{HuangAAAI2024}, and VecGAN++~\cite{DalvaTPAMI2023}, as shown in Table~\ref{Table 0}.

\subsection{Training Details and Evaluation Metrics}\label{sec4.2}

\vspace{0.5em}\noindent {\textbf{Training Details.} In our model, the denoising UNet $\epsilon$ and the style modulation encoder $F$ (excluding the style modulation module) are initialized with pre-trained parameters from the UNet network and semantic encoder of DiffAE~\cite{PreechakulCVPR2022}, respectively. These pre-trained parameters remain frozen, while the remaining components of the model are fine-tuned using the Adam optimizer. The learning rates are set to $0.0001$ for the style modulation module and $E$, and $0.000001$ for $M$. All input images are resized to $256 \times 256$, and the batch size is fixed at $16$. Training is completed after $60$ epochs. The source code for the proposed model, along with additional comparative results, is publicly available online\footnote{https://github.com/WeMiHuang/LatRef-Diff}.}

\vspace{0.5em}\noindent  \textbf{Evaluation Metrics.}
The quality of edited images and the success rate of editing are two pivotal metrics in both style manipulation and attribute editing. Consistent with prior methods, we use the Frechet Inception Distance (\textbf{FID})~\cite{HeuselNIPS2017} to measure the statistical distance between edited and real images. This metric evaluates not only the quality of the edited images but also the model's ability to generate diverse images.
In Setting \#1, the FID for the attributes $Bangs$ and $Eyeglasses$ is reported. For each attribute, we generate five images in different styles for each test image and calculate the FID between these edited images and the real images with that attribute in the training set.
In Setting \#2, a similar approach is used for the attributes $Smile$, $Young$, and $Male$. The key difference is that only one edited image is generated for each test image in this setting.
To evaluate the success rate of editing, we use the widely adopted attribute editing accuracy metric (\textbf{Acc})\cite{DalvaTPAMI2023,HuangAAAI2024}. This involves using an attribute classifier based on ResNet-18\cite{HeCVPR2016}, trained on the first $27,000$ images of CelebA-HQ. The classifier achieves an average accuracy of $95.0\%$ on all attributes in the remaining $3,000$ images of CelebA-HQ.

\begin{table*}[ht]
\caption{Quantitative Results of Style Manipulation in Latent-Guided Manner under setting \#1}\label{Table 1}
      \centering
      \makegapedcells
      \setcellgapes{2pt}
      \setlength\tabcolsep{10pt}
      \begin{tabular}{c || c | c | c| c| c | c | c| c|c| c}
        \hline
        \multirow{2}{*}{Method} &\multicolumn{2}{c|}{+ Bangs}&\multicolumn{2}{c|}{- Bangs}& \multicolumn{2}{c|}{+ Eleglasses}& \multicolumn{2}{c|}{- Eleglasses}&\multicolumn{2}{c}{Average}\\
        \cline{2-11}
        &$\downarrow{}$FID &$\uparrow{}$Acc &$\downarrow{}$FID &$\uparrow{}$Acc &$\downarrow{}$FID &$\uparrow{}$Acc &$\downarrow{}$FID &$\uparrow{}$Acc&$\downarrow{}$FID &$\uparrow{}$Acc\\
        \hline
        \hline
        HiSD\cite{LiCVPR2021}&17.18&85.99&39.71&92.25&62.17 &80.39& 95.99 &99.33&53.76&89.49\\
        \hline
        VecGAN++\cite{DalvaTPAMI2023}&17.12&84.98&36.88&90.21&-&-&-&-&-&-\\
        \hline
        SDGAN~\cite{HuangAAAI2024}&15.69&\textbf{90.67}&38.69&97.67&\textbf{49.39}&98.52&78.99&95.33&45.69&95.54\\
        \hline
        BSGAN~\cite{ren2025facial}&16.69&84.77&38.19&91.01&56.94&90.10&81.05&90.66&48.21&89.13\\
        \hline
        LatRef-Diff &\textbf{15.53}&90.16&\textbf{35.68}&\textbf{98.16}&50.02&\textbf{98.57}&\textbf{71.88}&\textbf{99.51}&\textbf{43.27}&\textbf{96.60}\\
        \hline
      \end{tabular}

\end{table*}

\begin{table*}[ht]
\caption{Quantitative Results of Style Manipulation in reference-guided manner under setting \#1}\label{Table 2}
      \centering
      \makegapedcells
      \setcellgapes{2pt}
      \setlength\tabcolsep{10pt}
      \begin{tabular}{c || c | c | c| c| c | c | c| c|c| c}
        \hline
        \multirow{2}{*}{Method} &\multicolumn{2}{c|}{+ Bangs}&\multicolumn{2}{c|}{- Bangs}& \multicolumn{2}{c|}{+ Eleglasses}& \multicolumn{2}{c|}{- Eleglasses}&\multicolumn{2}{c}{Average}\\
        \cline{2-11}
        &$\downarrow{}$FID &$\uparrow{}$Acc &$\downarrow{}$FID &$\uparrow{}$Acc &$\downarrow{}$FID &$\uparrow{}$Acc &$\downarrow{}$FID &$\uparrow{}$Acc&$\downarrow{}$FID &$\uparrow{}$Acc\\
        \hline
        \hline
        ELEGANT\cite{XiaoECCV2018}&28.16&72.88 &52.11&85.28&91.10&68.27&107.08&98.66&69.61&81.27\\
        \hline
        HiSD\cite{LiCVPR2021}&17.53&76.32&38.73&89.33& 56.64 &83.82& 89.97 &95.11&50.71&86.14\\
        \hline
        VecGAN++\cite{DalvaTPAMI2023} &16.37&78.99&37.44&91.65&-&-&-&-&-&-\\
        \hline
        SDGAN~\cite{HuangAAAI2024} &16.27&86.27&38.62&90.52&\textbf{51.45}&87.09&75.17&95.02&45.37&89.70\\
        \hline
        BSGAN~\cite{ren2025facial}&17.29&84.56&38.33&91.15&55.38&87.33&80.63&91.66&47.90&88.67\\
        \hline
        LatRef-Diff &\textbf{15.96}&\textbf{90.98}&\textbf{35.67}&\textbf{97.34}&52.16&\textbf{87.57}&\textbf{73.81}&\textbf{99.94}&\textbf{44.40}&\textbf{93.96}\\
        \hline
      \end{tabular}

\end{table*}


\subsection{Comparison of Style Manipulation}\label{sec4.3}
In this section, we evaluate our method's performance in style manipulation from both latent-guided and reference-guided perspectives, using qualitative and quantitative analyses. To provide a more comprehensive evaluation, we organize the comparison into two parts: (1) against face-attribute editing baselines, and (2) against recent text-to-image models.

\subsubsection{Comparison with Face-Attribute Editing Baselines}\label{sec4.3.1}
For this part, we compare our method with five advanced face-editing approaches. Notably, VecGAN++ does not support the addition or removal of eyeglasses; therefore, results for this attribute are excluded from both qualitative and quantitative evaluations. The qualitative comparison results are shown in Fig.~\ref{fig_4} and Fig.~\ref{fig_5}. From these figures, we conduct a comparative analysis on three key dimensions: unrelated attribute preservation, style manipulation effectiveness, and realism.

\begin{enumerate}
\item \textbf{Unrelated Attribute Preservation}:
Taking the addition of bangs as an example, ELEGANT, HiSD, VecGAN++, SDGAN, and BSGAN often introduce changes to regions unrelated to the bangs, such as the background, clothing, ears, mouth, and hat, leading to blurred or unnatural effects. In some cases, VecGAN++ even manipulates the male attribute incorrectly. For the eyeglasses editing task, ELEGANT produces substantial unintended modifications to the overall facial image.

\item \textbf{Style Manipulation Effectiveness}:
ELEGANT and VecGAN++ demonstrate limited capability in style manipulation when adding bangs, as they fail to effectively capture and transfer the bang style from the reference image, leading to uniform and monotonous results. Similar shortcomings are observed when removing bangs, with the generated hairstyles frequently failing to align with the reference image.

\item \textbf{Realism}:
ELEGANT, HiSD, SDGAN, and BSGAN struggle to produce realistic hair texture and detail when generating bangs. When creating eyeglasses, the shapes appear distorted, and lens reflections are poorly rendered, as shown in the second and third columns of Fig.~\ref{fig_5}(a). Moreover, when removing eyeglasses, the generated eyes lack realism, undermining the credibility of the edits.
\end{enumerate}

In contrast, our method accurately manipulates the style of the target attribute based on the reference image while preserving unrelated regions. Across all three evaluation dimensions, our approach significantly outperforms existing methods, achieving superior performance in unrelated attribute preservation, style manipulation effectiveness, and realism.

\begin{figure*}[!ht]
\centering
\includegraphics[width=6.8in]{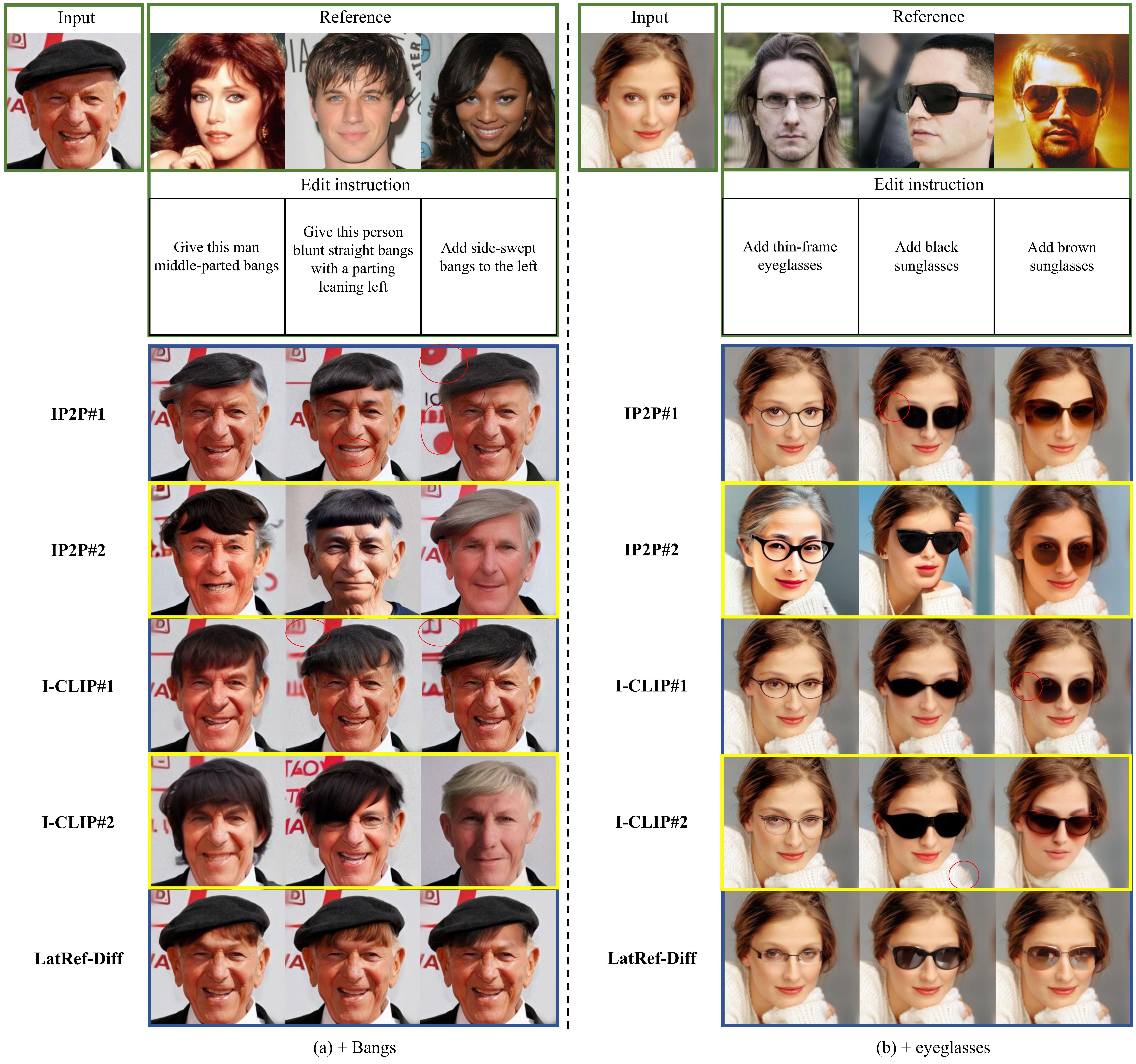}
\caption{Qualitative Results on Style Manipulation under setting \#1: Reference-Guided vs. Text-Guided, including (a) + Bangs and (b) + Eyeglasses.}
\label{fig_5_5}
\end{figure*}

\begin{figure*}[!ht]
\centering
\includegraphics[width=7.1in]{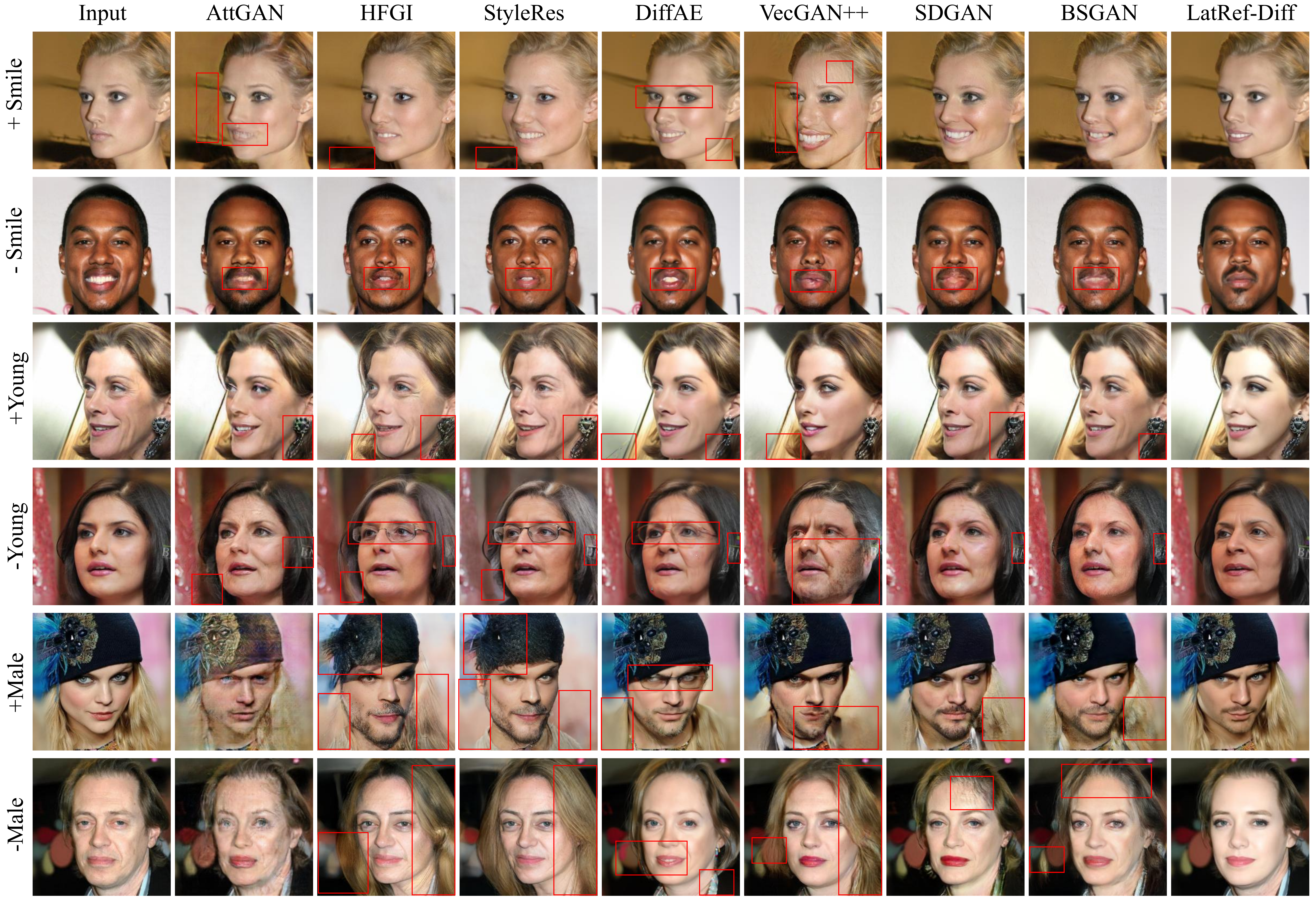}
\caption{Qualitative results of  attribute editing under setting \#2, including +Smile, -Smile, +Young, -Young, +Male, and -Male.}
\label{fig_6}
\end{figure*}

The quantitative results for latent-guided and reference-guided evaluations are presented in Tables~\ref{Table 1} and \ref{Table 2}, respectively. A detailed analysis of these tables reveals that our approach consistently outperforms other methods across nearly all attributes.  Notably, as shown in Table~\ref{Table 1}, our method achieves an average FID of $43.27$, marking a substantial improvement over the state-of-the-art SDGAN with a reduction of $2.42$. Furthermore, our approach achieves the highest mean accuracy rate of $96.60\%$, surpassing all competing methods.  In Table~\ref{Table 2}, our method continues to exhibit exceptional performance, particularly in terms of \textbf{Acc}. For example, the \textbf{Acc} for ``- Eyeglasses'' achieves an impressive $99.94\%$, setting a new benchmark. The overall average accuracy improves to $93.96\%$, reflecting a gain of $4.26\%$ compared to the closest competitor.  These results collectively underscore the superiority of our method in {style manipulation}, demonstrating its effectiveness and reliability across different evaluation scenarios.

\subsubsection{Comparison with Text-to-Image Models}
In recent years, large-scale text-to-image models have demonstrated remarkable advances in semantic generation and image synthesis. Nevertheless, these models generally lack straightforward editing mechanisms and often provide limited control over specific semantic regions of an input image. Even minor modifications to a text prompt can produce drastically different outputs, making them unreliable for fine-grained manipulation. To overcome these limitations, instruction-based editing methods trained on paired images with corresponding editing instructions have emerged, offering significantly stronger controllability for local editing. Considering the strict requirement in face attribute editing that unrelated regions must remain unchanged, we adopt two representative instruction-based approaches, IP2P~\cite{brooks2023instructpix2pix} and I-CLIP~\cite{chen2025instruct}, as extended baselines.

Both IP2P and I-CLIP enable flexible modifications of source images through editing instructions and incorporate two control scales, $s_T$ and $s_I$. Increasing $s_I$ makes the edited image closer to the input, while increasing $s_T$ amplifies the editing strength. To evaluate the style manipulation capability of these methods, we design editing instructions based on the reference image styles and then compare the manipulation results. For a fair comparison, we carefully tune the editing instructions, $s_T$, $s_I$, and the random sampling variable $x_T$ to obtain the best possible outcomes, denoted as IP2P\#1 and I-CLIP\#1. To further assess stability, we keep the instructions and scales fixed while varying $x_T$, resulting in IP2P\#2 and I-CLIP\#2. The qualitative results are shown in Fig.~\ref{fig_5_5}. From these results, we observe:
\begin{itemize}
    \item Although IP2P and I-CLIP can manipulate styles through text prompts, they often fail to accurately transfer the reference style, as fine-grained style details are difficult to capture using language alone.
    \item Comparisons between IP2P\#1 vs. IP2P\#2 and I-CLIP\#1 vs. I-CLIP\#2 reveal a strong sensitivity to the random sampling $x_T$, which leads to large variations in editing results and increases the overall complexity of the editing process.
    \item Even with extensive tuning of instructions and parameters ($s_T$, $s_I$, and $x_T$) to obtain the best possible results (IP2P\#1 and I-CLIP\#1), the results remain unsatisfactory. For instance, when adding bangs, the models sometimes fail to recognize hats and may incorrectly alter the mouth or background; when adding eyeglasses, the generated glasses are often incomplete.
    \item In contrast, our method consistently produces realistic edits that faithfully reflect the intended style without the need for laborious parameter tuning, yielding more stable and visually convincing results.
\end{itemize}

Finally, we emphasize that IP2P and I-CLIP are extremely sensitive to editing instructions, parameter settings, and random sampling. Achieving satisfactory editing results requires labor-intensive tuning for each individual image, which makes large-scale evaluation impractical. Consequently, we do not provide quantitative results for these methods.

\subsection{Comparison of Attribute Editing}\label{sec4.4}
In this section, we evaluate the attribute editing performance of the proposed work and other related  methods. Fig.~\ref{fig_6} showcases the qualitative results of this comparison. Our evaluation emphasizes three key aspects: unrelated attribute preservation, attribute editing effectiveness, and realism.
\begin{enumerate}
\item\textbf{Unrelated Attribute Preservation}: A common limitation among the compared methods is their tendency to unintentionally alter unrelated regions, such as the background, ornaments (e.g., earrings, headgear), hair, and clothing, which are highlighted with red boxes in the figure. In addition, DiffAE  modifies eye gaze when adding a smile (+Smile) and alters the mouth shape when removing male attributes (-Male).

\item \textbf{Attribute Editing Effectiveness}: When removing a smile (-Smile), methods like HFGI, StyleRes, and DiffAE struggle to fully close the mouth, leaving traces of a smile. Similarly, when attempting to age a subject (-Young), these methods often inadvertently modify unrelated features such as eyeglasses, while VecGAN++ introduces incorrect changes to gender attributes. Furthermore, when making subjects look younger (+Young), the outputs from HFGI and StyleRes still exhibit signs of aging, such as wrinkles.

\item \textbf{Realism}: Some methods introduce visible artifacts that compromise realism. For instance, AttGAN’s outputs for adding or removing a smile (+/-Smile) introduce noticeable artifacts around the mouth and significant noise during gender transitions (+/-Male). VecGAN++ distorts facial contours during the +Smile'' operation and produces inconsistent mouth features in the -Smile'' operation. Similarly, both SDGAN and BSGAN generate  prominent artifacts when removing a smile (-Smile).
\end{enumerate}

\begin{table*}[!ht]
\caption{Quantitative results of attribute editing under setting \#2}\label{Table 3}
      \centering
      \makegapedcells
      \setcellgapes{2pt}
      \setlength\tabcolsep{6pt}
      \begin{tabular}{c || c | c | c| c| c | c | c| c|c| c|c| c|c| c}
        \hline
        \multirow{2}{*}{Method} &\multicolumn{2}{c|}{+ Smile}&\multicolumn{2}{c|}{- Smile}& \multicolumn{2}{c|}{+ Young}& \multicolumn{2}{c|}{- Young}& \multicolumn{2}{c|}{+ Male}& \multicolumn{2}{c|}{- Male}&\multicolumn{2}{c}{Average}\\
        \cline{2-15}
        &$\downarrow{}$FID &$\uparrow{}$Acc &$\downarrow{}$FID &$\uparrow{}$Acc &$\downarrow{}$FID &$\uparrow{}$Acc &$\downarrow{}$FID &$\uparrow{}$Acc&$\downarrow{}$FID &$\uparrow{}$Acc&$\downarrow{}$FID &$\uparrow{}$Acc&$\downarrow{}$FID &$\uparrow{}$Acc\\
        \hline
        \hline
         AttGAN\cite{HeTIP2019} &25.22&94.35&28.56&97.05&64.83&77.51&36.73&86.98&70.68&97.98&91.02&78.80&52.84&88.77\\
        \hline
         HFGI\cite{WangCVPR2022}&29.59&83.77&36.69&91.23&54.37&80.39&51.81&81.34&67.33&97.37&52.17&88.68&48.66&87.13\\
         \hline
         StyleRes\cite{PehlivanCVPR2023}&23.05&94.61&25.81&96.71&50.99&87.92&32.36&82.84&63.38&92.13&49.20&94.57&40.79&91.46\\
         \hline
         DiffAE\cite{PreechakulCVPR2022}&20.42&94.21&25.50&90.86&47.43&78.59&26.27&\textbf{92.22}&42.98&96.76&41.77&97.01&34.06&91.60\\
         \hline
         SDGAN~\cite{HuangAAAI2024}&21.80&88.38&26.74&86.85&40.51&82.56&32.23&74.86&45.89&93.65&34.37&95.65&33.59&86.99\\
         \hline
         VecGAN++\cite{DalvaTPAMI2023}&20.60&97.88&26.13&95.11&36.86&93.44&26.61&80.38&46.97&86.95&\textbf{33.68}&85.97&31.80&89.95\\
         \hline
         BSGAN\cite{DalvaTPAMI2023}&19.31&80.55&26.22&89.43&42.23&80.96&33.69&75.05&44.79&85.11&33.79&89.61&33.34&83.45\\
         \hline
         LatRef-Diff &\textbf{16.81}&\textbf{99.29}&\textbf{22.94}&\textbf{97.55}&\textbf{36.09}&\textbf{97.90}&\textbf{25.84}&89.62&\textbf{40.47}&\textbf{98.35}&38.70&\textbf{99.54}&\textbf{30.14}&\textbf{97.04}\\
        \hline
      \end{tabular}
\end{table*}

These evaluations indicate that the compared methods frequently struggle to produce satisfactory editing results. In contrast, our approach demonstrates superior performance by preserving unrelated attributes, achieving precise editing of target attributes, and generating highly realistic results.

The quantitative results are detailed in Table~\ref{Table 3}, demonstrating that our method outperforms existing approaches across nearly all attributes. Notably, our method achieves an average FID of $30.14$, indicating a significant improvement in image quality over the previous leading method, VecGAN++, which has an average FID of $31.80$. In terms of Acc, our approach achieves an average of $97.04$\%, surpassing DiffAE, the previous best performer, by $5.44$ percentage points (DiffAE: $91.60$\%). These results highlight the substantial advancements our method brings to precise attribute editing.  In detailed analyses of specific attributes, our method also demonstrates clear superiority. For example, in the case of the ``Smile'' attribute, our approach excels in both addition and removal tasks. In the ``+Smile'' task, our FID score is $16.81$, improving upon BSGAN's best score by $2.5$ points, while our accuracy reaches $99.29$\%, surpassing the previous best by VecGAN++ by $1.41$ percentage points. In the ``-Smiling'' task, our FID score is $22.94$, an improvement over DiffAE's result by $2.56$ points, and the accuracy climbs to $97.55$\%, outperforming the top result by AttGAN. Overall, these results validate the robustness, adaptability, and superiority of our approach in handling a wide range of attribute editing challenges.

\subsection{Ablation Study}\label{sec4.5}
\begin{figure*}[!ht]
\centering
\includegraphics[width=7.0in]{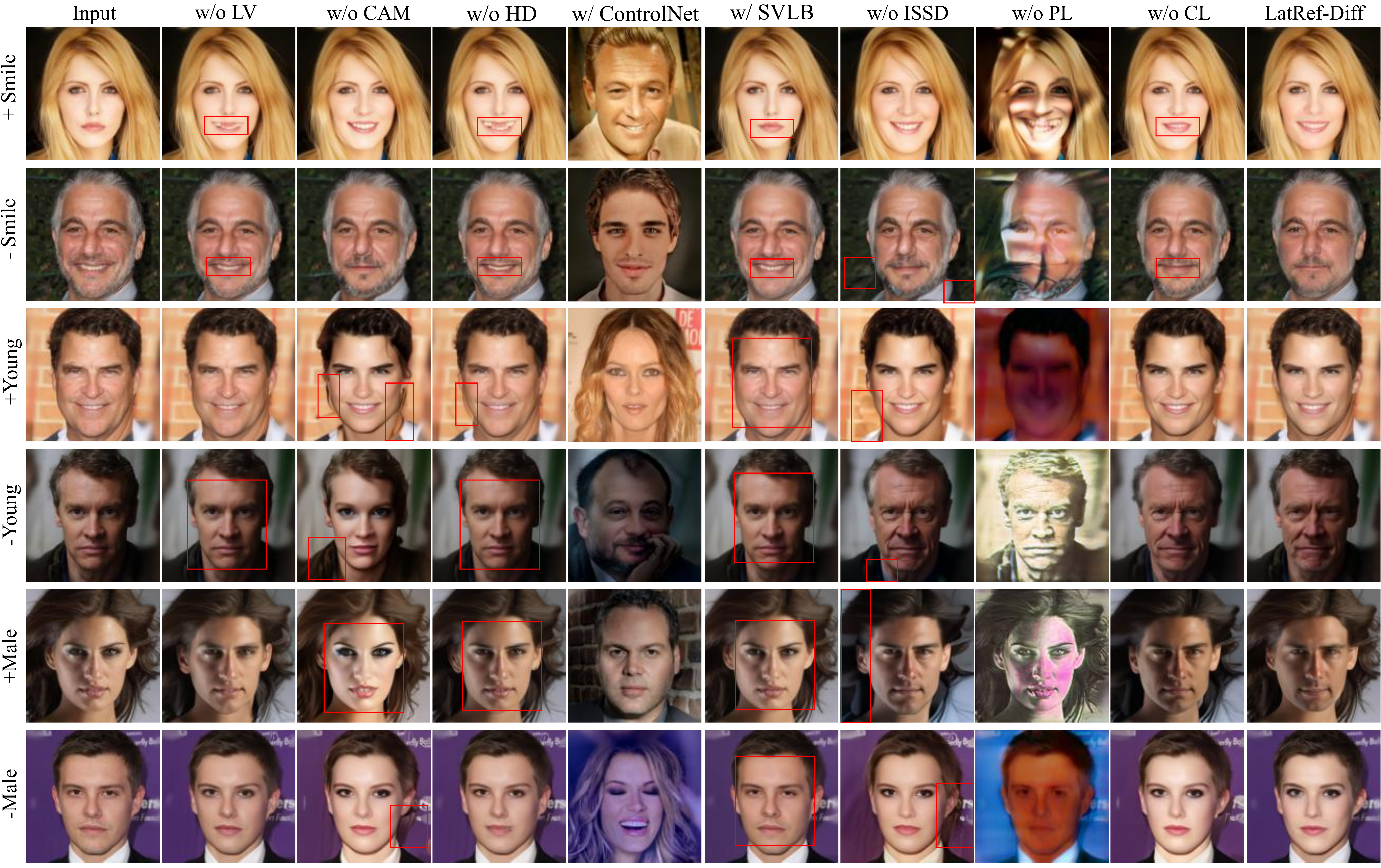}
\caption{Qualitative results of the ablation study, including +Smile, -Smile, +Young, -Young, +Male, and -Male.}
\label{fig_7}
\end{figure*}

As discussed in Section III, our approach incorporates two critical components: the style modulation module and the forward-backward consistency training strategy. This section verifies their effectiveness through ablation studies.
The qualitative and quantitative results are depicted in Fig.~\ref{fig_7} and summarized in Table \ref{Table 4}, respectively. Overall, LatRef-Diff exhibits superior image editing quality in qualitative evaluations and consistently surpasses alternative configurations in nearly all quantitative metrics. Subsequently, we will analyze the effectiveness of these two components based on the results from Fig.~\ref{fig_7} and Table \ref{Table 4}.

\subsubsection{Effectiveness of Style Modulation Module}\label{sec4.5.1}
The style modulation module aims to significantly improve the editing images quality and editing accuracy through learnable vectors, cross-attention mechanisms, and a hierarchical design. To assess the individual contributions of these key elements, we systematically remove each one in turn and evaluate the impact on the performance of LatRef-Diff.

\vspace{0.5em}\noindent  \textbf{Learnable Vectors (LV).} Learnable vectors are introduced as additional conditional inputs for AdaIN, aiming to enhance the accuracy of attribute editing. To assess the impact of this element, we remove the learnable vector, creating a variant denoted as ``w/o LV''.
Qualitatively, its absence leads to noticeable artifacts noticeable artifacts in the ``+Smile'' manipulation and complete failures in the ``-Smile'' and ``-Young'' manipulations.
The quantitative results further highlight the importance of LV. Compared to LatRef-Diff, the average FID score for ``w/o LV'' increases by $16.83$ points, indicating a significant decline in image quality.
Additionally, ``w/o LV'' exhibits significant drops in accuracy (Acc), with reductions of $58.94$\%, $68.37$\%, and $48.61$\%  for ``-Smile'', ``-Young'', and average Acc, respectively.

\vspace{0.5em}\noindent  \textbf{Cross-Attention Mechanisms (CAM).}
Cross-attention mechanisms are employed to refine local details and adjust global structures, improving the quality of edited images.  To evaluate the contribution of this element, we remove the cross-attention mechanisms, resulting in a variant referred to as ``w/o CAM''.
Qualitatively, the absence of CAM introduces noticeable artifacts and inconsistencies. For instance, ``w/o CAM'' causes visible distortions around facial contours in the ``+Young'' manipulation, improperly alters clothing in the ``-Young'' task, results in severe facial feature distortions in the ``+Male'' task, and modifies the background in the ``-Male'' manipulation.
Quantitatively, the removal of CAM leads to performance degradation in both FID and Acc metrics. Specifically, the average FID score increases by $1.04$ points, and the average Acc decreases by $2.72$\% points compared to LatRef-Diff.

\vspace{0.5em}\noindent  \textbf{Hierarchical Design (HD).}
The hierarchical design mitigates interference between different attributes by assigning a dedicated style modulation unit for each tag and a learnable vector for each attribute. To evaluate the impact of this design, we introduce a variant referred to as ``w/o HD'', where all attributes share a single style modulation unit and a learnable vector.
Qualitatively, ``w/o HD''  produces noticeable artifacts in the manipulations of ``+Smile'' and ``+Young'' and fails entirely in ``-Smile'', ``-Young'', and ``+Male''.
Quantitatively, this variant shows significant performance deterioration across all attributes in both FID and Acc metrics compared to LatRef-Diff. For instance, the average FID worsens by $18.83$ points, and the average Acc drops by $51.88$\%. The disparity in Acc between opposing attribute for tag is particularly pronounced, with differences of $68.95$\% for ``+/-Smile'', $53.86$\% for ``+/-Young'' and $58.35$\% points for ``+/-Male''.

\vspace{0.5em}\noindent \textbf{ControlNet.}
To further assess the effectiveness of our style modulation mechanism in fine-grained face editing, we replace the modules $M$, $E$, and $F$ in our framework with the recently proposed ControlNet\cite{zhang2023adding}, which is designed to improve the editing controllability of text-to-image diffusion models. This variant is denoted as ``w/ ControlNet''. From the qualitative results, we observe that although ``w/ ControlNet'' is able to generate images with the target attributes, it entirely alters the content of the input image and essentially degenerates the editing task into conditional image generation, i.e., generating a new face image with the desired attributes. Since this falls outside the scope of facial attribute editing, we exclude it from quantitative evaluation.

\begin{table*}[ht]
\caption{Quantitative results of the ablation study}\label{Table 4}
      \centering
      \makegapedcells
      \setcellgapes{2pt}
      \setlength\tabcolsep{6pt}
      \begin{tabular}{c || c | c | c| c| c | c | c| c|c| c|c| c|c| c}
        \hline
        \multirow{2}{*}{Method} &\multicolumn{2}{c|}{+ Smile}&\multicolumn{2}{c|}{- Smile}& \multicolumn{2}{c|}{+ Young}& \multicolumn{2}{c|}{- Young}& \multicolumn{2}{c|}{+ Male}& \multicolumn{2}{c|}{- Male}&\multicolumn{2}{c}{Average}\\
        \cline{2-15}
        &$\downarrow{}$FID &$\uparrow{}$Acc &$\downarrow{}$FID &$\uparrow{}$Acc &$\downarrow{}$FID &$\uparrow{}$Acc &$\downarrow{}$FID &$\uparrow{}$Acc&$\downarrow{}$FID &$\uparrow{}$Acc&$\downarrow{}$FID &$\uparrow{}$Acc&$\downarrow{}$FID &$\uparrow{}$Acc\\
        \hline
        \hline
         w/o LV &21.10&48.45&25.22&38.61&57.15&42.66&49.90&21.25&68.18&57.31&60.30&82.33&46.97&48.43\\

         w/o CAM &17.95&98.39&24.50&97.31&36.74&\textbf{98.20}&27.73&74.90&41.16&98.26&39.05&98.90&31.18&94.32\\

         w/o HD &19.29&80.01&24.53&11.06&48.11&68.26&47.14&14.40&91.21&19.45&63.59&77.80&48.97&45.16\\
         \hline
         \hline
         w/ SVLB &25.44&2.76&26.87&4.47&60.66&3.27&56.28&0.82&125.14&0.67&112.06&3.89&67.74&2.64\\
         w/o ISSD &17.96&97.80&25.42&92.73&36.44&97.90&26.74&84.41&\textbf{40.42}&97.45&\textbf{38.25}&98.72&30.87&94.83\\
         w/o PL &48.81&25.57&48.50&37.07&129.57&48.95&94.68&18.82&115.74&10.77&168.68&0.97&100.99&23.69\\
         w/o CL&18.70&82.96&22.97&79.39&36.77&93.11&26.65&48.84&41.89&97.71&44.10&98.91&31.84&83.48\\
         \hline
         LatRef-Diff &\textbf{16.81}&\textbf{99.29}&\textbf{22.94}&\textbf{97.55}&\textbf{36.09}&97.90&\textbf{25.84}&\textbf{89.62}&40.47&\textbf{98.35}&38.70&\textbf{99.54}&\textbf{30.14}&\textbf{97.04}\\
        \hline
      \end{tabular}
\end{table*}

\subsubsection{Effectiveness of Forward-Backward Consistency Training Strategy}\label{sec4.5.2}
The forward-backward consistency training strategy tackles the challenge of training without paired images by utilizing image-specific semantic direction, perceptual loss, and classification loss. Below, we will validate the effectiveness of our training strategy by replacing or directly removing key elements of our approach using related approaches.

\vspace{0.5em}\noindent \textbf{Forward-Backward Consistency Training Strategy (FBCTS).}
To validate the effectiveness of our FBCTS, we introduce a variant model that is trained using a simplified version of the variational lower bound (SVLB) commonly employed in diffusion models, $||\epsilon_\theta(x_{t,(i,j)}, t,c_{i,j}) - \epsilon||_2^2$, referred to as ``w/ SVLB''. Qualitatively, the output of ``w/ SVLB'' merely reconstructs the input, lacking the ability to edit various attributes. Qualitative analysis further supports this observation, where ``w/ SVLB'' achieves an average Acc of only $2.6$\%, demonstrating minimal attribute editing ability. The primary reason lies in the training process: $F$ bypasses the style $s_{i,j}$ and learns a  trivial mapping $c_{i,j} = F(x_{i,j}, \cdot)$, as discussed in Section III-D. Consequently, the style modulation module fails to learn the intended attribute editing operations as expected.

\vspace{0.5em}\noindent \textbf{Image-Specific Semantic Direction (ISSD).}
We propose the image-specific semantic direction $d_{t}$ as a replacement for the original $d_{s}$ to address unintended modifications to non-target attributes during editing. For comparison, we introduce a variant referred to as ``w/o ISSD'', where $d_{s}$ is used in the forward-backward consistency training strategy to remove target attributes.
Qualitative results reveal that ``w/o ISSD'' leads to undesired changes in backgrounds and clothing during attribute manipulations such as ``+Smile'', ``+Young'', ``-Young'', and ``-Male''. Additionally, it alters non-target attributes, such as hair color, when editing ``+Male''.
Quantitatively, the variant ``w/o ISSD'' exhibits significant declines in both FID and Acc metrics for the``+Smile'', ``-Smile'', and ``-Young'' manipulations, with average Acc $2.21\%$ lower than that of LatRef-Diff.

\vspace{0.5em}\noindent  \textbf{Perceptual Loss (PL).}
Perceptual loss is introduced to maintain perceptual similarity between the edited image and the input image. To evaluate its impact, we remove the perceptual loss, creating a variant referred to as ``w/o PL''.
Qualitatively, the absence of perceptual loss leads to noticeable artifacts in the edited images and a failure to preserve the content of the input image.
Qualitative analysis shows that ``w/o PL'' results in a significant decline in both image quality and editing accuracy, with average FID and Acc worsening by $70.85$ points and $73.35$\% , respectively, compared to LatRef-Diff.

\vspace{0.5em}\noindent  \textbf{Classification Loss (CL).}
Classification loss aims to ensure that the style modulation module effectively edits attributes to restore the target attributes. To evaluate its impact, we remove the classification loss, creating a variant referred to as ``w/o CL''.
Qualitatively, ``w/o CL'' struggles to accurately add or remove smiles, demonstrating limited attribute editing ability.
Quantitatively, the absence of classification loss results in a deterioration of the average FID by $1.70$ points compared to LatRef-Diff. Additionally, the Acc of  ``w/o CL'' decreases by $16.33$\%, $18.16$\%, and $40.78$\%  in ``+Smile'', ``-Smile'', and ``-Young'' manipulations, respectively, leading to a drop of $13.56$\% in average Acc.

\section{Conclusion}\label{sec5}
In this paper, we introduce  LatRef-Diff, a novel diffusion-based framework designed for facial attribute editing and style manipulation. Unlike previous diffusion models that rely on semantic directions, our method incorporates a style modulation mechanism based on style codes and a style modulation module. These style codes, generated from either randomly sampled Gaussian noise or reference images, are integrated into the target image through the style modulation module. This allows for both random and customized style manipulation while providing greater flexibility in attribute editing. The style modulation module leverages learnable vectors, cross-attention mechanisms, and a hierarchical design to ensure optimal performance. To further improve training stability and eliminate the need for paired images or adversarial loss, we propose a novel  forward-backward consistency training strategy. This strategy utilizes image-specific semantic directions alongside perceptual and classification losses, resulting in more robust model training. We validate the effectiveness of LatRef-Diff through extensive experiments on the CelebA-HQ dataset, where our framework outperforms existing methods in both precise style manipulation and high-quality attribute editing. Additionally, numerous ablation studies further highlight the significance of each model component, reinforcing the strengths of our approach.

Future work will explore two main directions: First, replacing the current MLP and CNN architectures with advanced models like transformers for improved style code extraction. Second, fine-tuning the entire LatRef-Diff framework, beyond the current mapper, extractor, and style modulation module, to further enhance editing quality.

\bibliographystyle{IEEEtran}
\bibliography{reference}

\begin{IEEEbiography}
	[{\includegraphics[width=1in,height=1.25in,clip,keepaspectratio]{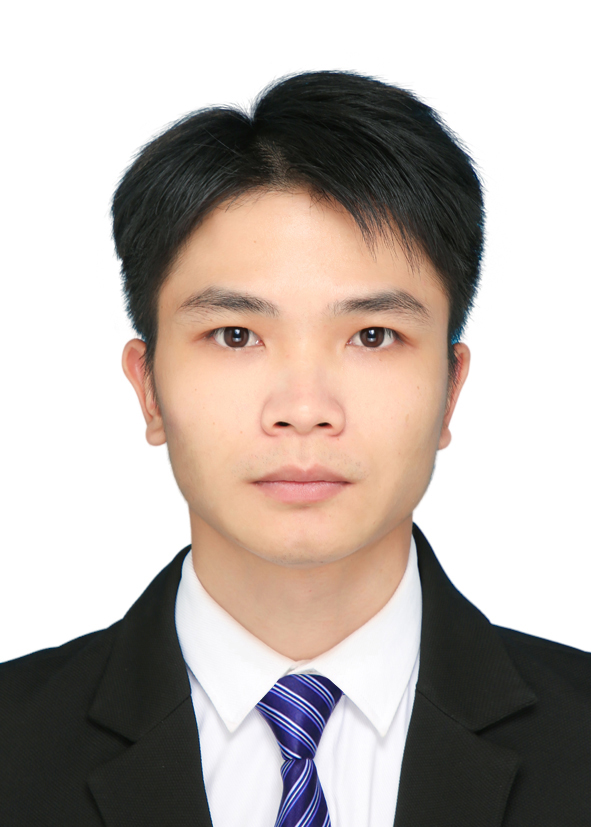}}]{Wenmin Huang}
	received the M.E. degree in intelligent science and technology from Tianjin Normal University, in 2021. He is currently working toward the Ph.D. degree at Sun Yat-sen University. His research interests include face attribute editing and computer vision.  
	He serves as a Program Committee for AAAI-2026 and a reviewer for IEEE Transactions on Circuits and Systems for Video Technology.
\end{IEEEbiography}

\begin{IEEEbiography}
	[{\includegraphics[width=1in,height=1.25in,clip,keepaspectratio]{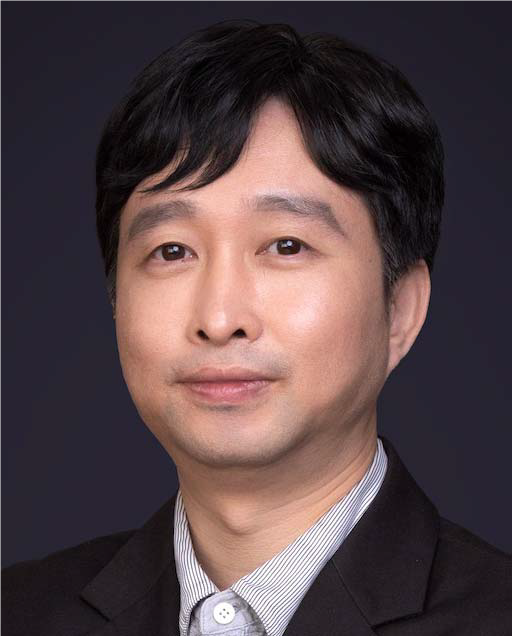}}]{Weiqi Luo} (Senior Member, IEEE) received his Ph.D. degree in 2008 from Sun Yat-sen University, Guangzhou, China. He is currently a Full Professor in the School of Computer Science and Engineering at Sun Yat-sen University. His research interests focus on digital multimedia forensics, steganography, and steganalysis. He serves as an Associate Editor for IEEE Transactions on Circuits and Systems for Video Technology, Information Forensics and Security, and Dependable and Secure Computing.
\end{IEEEbiography}	

\begin{IEEEbiography}
	[{\includegraphics[width=1in,height=1.25in,clip,keepaspectratio]{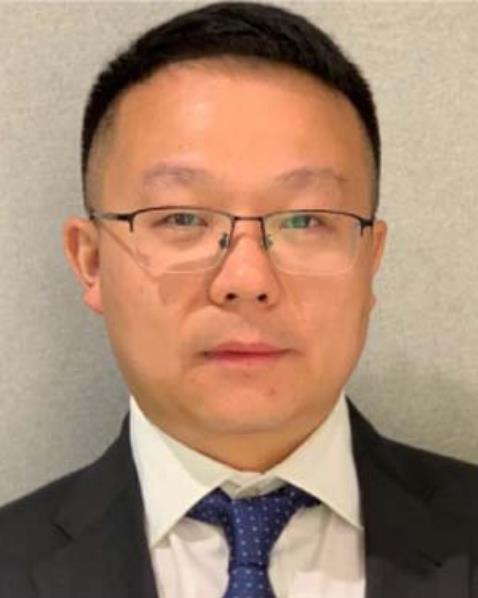}}]{Xiaochun Cao} (Senior Member, IEEE) received the B.E. and M.E. degrees in computer science from Beihang University (BUAA), China, and the Ph.D. degree in computer science from the University of Central Florida, USA. He is currently a Professor and the Dean of the School of Cyber Science and Technology, Shenzhen Campus of Sun Yat-sen University. His dissertation nominated for the university level Outstanding Dissertation Award. After graduation, he spent about three years with ObjectVideo Inc., as a Research Scientist. From 2008 to 2012, he was a Professor with Tianjin University. Before joining SYSU, he was a Professor with the Institute of Information Engineering, Chinese Academy of Sciences. He has authored or co-authored more than 200 journals and conference papers. In 2004 and 2010, he was a recipient of the Piero Zamperoni Best Student Paper Award at the International Conference on Pattern Recognition. He is on the editorial boards of IEEE Transactions on Pattern Analysis and Machine Intelligence and IEEE Transactions on Image Processing and was on the editorial boards of IEEE Transactions on Circuits and Systems for Video Technology and IEEE Transactions on Multimedia.
\end{IEEEbiography}	

\begin{IEEEbiography}
	[{\includegraphics[width=1in,height=1.25in,clip,keepaspectratio]{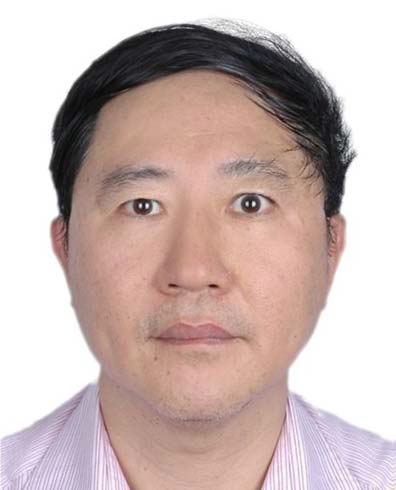}}]{Jiwu Huang}
	(Fellow, IEEE) received the B.S. degree from Xidian University, Xi’an, China, in 1982, the M.S. degree from Tsinghua University, Beijing, China, in 1987, and the Ph.D. degree from the Institute of Automation, Chinese Academy of Sciences, Beijing, in 1998. He is currently a Chief Professor with the Faculty of Engineering, Shenzhen MSU-BIT University, Shenzhen, 518116, China. He has coauthored more than 350 papers in important journals and conferences, with Google citations of more than 25000 and an H-index of 84. His current research interests include multimedia forensics and security.
\end{IEEEbiography}

\end{document}